\documentclass{article}

\usepackage[preprint]{neurips_2024}
\usepackage{amsmath}
\usepackage{graphicx}
\usepackage{svg}
\usepackage{subcaption}
\usepackage{changepage}
\usepackage{float}

\usepackage[utf8]{inputenc} 
\usepackage[T1]{fontenc} 
\usepackage{hyperref} 
\usepackage{url}
\usepackage{booktabs} 
\usepackage{amsfonts} 
\usepackage{nicefrac} 
\usepackage{microtype} 
\usepackage{xcolor} 

\title{Expanded Gating Ranges \\ Improve Activation Functions}

\author{%
 Allen Hao Huang \\
 Machine Learning and Optimization Laboratory \\
 EPFL \\
}

\begin{document}

\maketitle

\begin{abstract}

Activation functions are core components of all deep learning architectures. Currently, the most popular activation functions are smooth ReLU variants like GELU and SiLU. These are self-gated activation functions where the range of the gating function is between zero and one. In this paper, we explore the viability of using arctan as a gating mechanism. A self-gated activation function that uses arctan as its gating function has a monotonically increasing first derivative. To make this activation function competitive, it is necessary to introduce a trainable parameter for every MLP block to expand the range of the gating function beyond zero and one. We find that this technique also improves existing self-gated activation functions. We conduct an empirical evaluation of Expanded ArcTan Linear Unit (xATLU), Expanded GELU (xGELU), and Expanded SiLU (xSiLU) and show that they outperform existing activation functions within a transformer architecture. Additionally, expanded gating ranges show promising results in improving first-order Gated Linear Units (GLU).
\end{abstract}

\section{Introduction}

Activation functions are crucial for introducing non-linearities in deep neural networks \citep{goodfellow2016deep}. Without them, neural networks would essentially function as linear models, unable to capture complex patterns and relationships in data. Early neural networks employed activation functions that produced outputs within a bounded range, such as the binary threshold unit \citep{mcculloch1943logical}, logistic sigmoid, and hyperbolic tangent \citep{rumelhart1986learning}. However, using these activation functions in deep networks often led to the vanishing gradient problem, which adversely affected performance.

The Rectified Linear Unit (ReLU) \citep{glorot2011deep, agarap2018deep} became popular by addressing the limitations of traditional bounded activation functions. ReLU is computed by applying the binary threshold unit to the input and then multiplying the result by the input. It avoids the vanishing gradient problem by maintaining a constant, non-zero gradient for positive inputs, leading to faster convergence and improved training efficiency. Smooth ReLU variants, such as the Gaussian Error Linear Unit (GELU) \citep{hendrycks2016gaussian} and Sigmoid Linear Unit (SiLU / Swish) \citep{hendrycks2016gaussian, ramachandran2017searching, elfwing2017sigmoidweighted}, further improved upon ReLU by introducing continuously differentiable functions to replace the binary threshold unit.

\textbf{Contributions.} In this work, our objective is to deepen the understanding of existing activation functions and introduce improvements. Our contributions are as follows:

\begin{itemize}
\item \textbf{Favorable Properties.} We identify the properties that make activation functions effective. Our findings challenge the conventional belief that ReLU-like properties are necessary for good performance.
\item \textbf{Viability of Arctan.} We demonstrate the potential of using arctan as a gating mechanism. Arctan is competitive with and can even outperform existing gating functions in multiple settings.
\item \textbf{Improving Self-Gated Activation Functions.} We demonstrate that expanded gating ranges can enhance self-gated activation functions. We propose Expanded ArcTan Linear Unit (xATLU), Expanded GELU (xGELU), and Expanded SiLU (xSiLU), and show that they can outperform GELU and SiLU.
\item \textbf{Improving First-Order Gated Linear Units.} We demonstrate that expanded gating ranges can enhance first-order GLU and help bridge their performance gap with second-order GLU. We propose the first-order Expanded ArcTan GLU (xATGLU), Expanded GEGLU (xGEGLU), and Expanded SwiGLU (xSwiGLU), and show that they can achieve competitive performance with second-order GEGLU and SwiGLU.
\end{itemize}

\section{Preliminaries}

\subsection{Self-Gated Activation Functions}

Incorporating a self-gating mechanism within the activation function enhances gradient flow during training, leading to more stable and efficient learning dynamics within neural networks. Activation functions such as ReLU, GELU, and SiLU can be classified as self-gated activation functions. Gating functions typically output values within a bounded range, usually between zero and one, and can be interpreted as being responsible for controlling the flow of information. Self-gated activation functions can be expressed in the form:
\begin{equation}
a(x) = g(x) \times x
\label{eq:1}
\end{equation}
Where $x$ is the input, $a(x)$ is the self-gated activation function, and $g(x)$ is the gating function. The gating functions and ranges of popular activation functions, such as ReLU, GELU, and SiLU, are listed in Table \ref{tab:activation-functions}.

\begin{table}[h]
\caption{\textbf{Self-Gated Activation Functions.} ReLU, GELU and SiLU are self-gated activation functions. ReLU uses the binary threshold unit $(x > 0)$ as its gating function, GELU uses the standard Gaussian Error CDF $\phi(x)$ as its gating function, and SiLU uses the logistic sigmoid $\sigma(x)$ as its gating function. All the gating functions have a range between zero and one.}
\label{tab:activation-functions}
\centering
\begin{tabular}{ccc}
\midrule
$a(x)$ & $g(x)$ & $g(x)$ range \\
\midrule
ReLU & $x > 0$ & $[0, 1]$ \\
GELU & $\phi(x)$ & $(0, 1)$ \\ 
SiLU & $\sigma(x)$ & $(0, 1)$ \\
\bottomrule
\end{tabular}
\end{table}
Traditionally, the search for activation functions in neural networks has relied heavily on trial and error, with researchers exploring various functions and evaluating their performance empirically \citep{lecun1998gradient, nair2010rectified, clevert2016fast}. In this paper, we aim to identify the favorable properties of existing activation functions to guide the design of new and improved activation functions.

\subsection{Potentially Favourable Properties}
\label{2.2-favorable-properties}

GELU and SiLU are currently the state-of-the-art activation functions. We identify the following shared properties that are potentially useful:
\begin{enumerate}
\item $g(x)$ is monotonically increasing.
\item $g(x)$ is continuously differentiable.
\item $g(x)$ has a range between zero and one.
\item $a(x) \rightarrow 0$ as $x \rightarrow -\infty$ and $a(x) \rightarrow x$ as $x \rightarrow \infty$.
\item $a(x)$ has gradients that can be below zero and above one.
\end{enumerate}
We assume properties 1 and 2 are beneficial and do not run experiments to test them. A monotonically increasing $g(x)$ ensures that more important information is preserved and propagated through the activation function. $g(x)$ is monotonically increasing by convention, although it is possible to construct functionally equivalent activation functions with $g(x)$ being monotonically decreasing. $g(x)$ being continuously differentiable ensures smoother gradient flow during backpropagation and explains why smooth ReLU variants like GELU and SiLU outperform ReLU.

We run experiments to test the benefits of properties 3, 4, and 5. Property 4 can also be interpreted as $a(x)$ being a ReLU-like function. It is important to note the distinction between properties 3 and 4: although property 4 implies property 3, the converse is not true. The first derivatives of both GELU and SiLU have two turning points, which allow for gradients below zero and above one.

\subsection{Rescaling Function Ranges}

Activation functions appear to require specific gating ranges to perform effectively. For instance, ReLU, GELU and SiLU all have a gating range between zero and one. New gating functions can be created by rescaling the range of bounded functions to this specific gating range. This approach has precedence in existing activation functions. For example, $\phi(x)$, the gating function of GELU, can be interpreted as the result of rescaling erf$(x/\sqrt{2})$ from the range of $(-1, 1)$ to $(0, 1)$. Similarly, $\sigma(x)$, the gating function of SiLU, can be interpreted as the result of rescaling tanh$(x)$ from the range of $(-1, 1)$ to $(0, 1)$.

To rescale the range of a function $f(x)$ from $(\text{min}_{\text{old}}, \text{max}_{\text{old}})$ to $(\text{min}_{\text{new}}, \text{max}_{\text{new}})$, we can use the following linear transformation:
\begin{equation}
f_{\text{new}}(x) = (f(x) - \text{min}_{\text{old}}) \left(\frac{{\text{max}_{\text{new}} - \text{min}_{\text{new}}}}{{\text{max}_{\text{old}} - \text{min}_{\text{old}}}}\right) + \text{min}_{\text{new}} 
\label{eq:2}
\end{equation}
This transformation adjusts the output range of the function $f(x)$ to the desired new range, facilitating the creation of new gating functions from bounded functions.

\subsection{Gated Linear Units}

Gated Linear Units (GLU) \citep{dauphin2017language} are activation functions defined as the component-wise product of two inputs, where one input is passed through a non-linearity. The key difference between GLU and self-gated activation functions is that self-gated activation functions compute the component-wise product of a single input with itself after it has been passed through a non-linearity.

We experiment with first-order GLU of the form:
\begin{equation}
a(x, y) = g(x) \times y
\label{eq:3}
\end{equation}

We experiment with second-order GLU of the form:
\begin{equation}
a(x, y) = g(x) \times x \times y
\label{eq:4}
\end{equation}

An example of a first-order GLU is the original GLU where $g(x)$ is $\sigma(x)$. Second-order GLU \citep{shazeer2020glu} introduce an additional multiplicative interaction compared to first-order GLU and tend to have better performance. Popular second-order GLU include ReGLU, where $g(x)$ is $x > 0$, GEGLU, where $g(x)$ is $\phi(x)$, and SwiGLU, where $g(x)$ is $\sigma(x)$.

\section{Methodology}

\subsection{ArcTan Linear Unit}

\begin{figure}[t!]
\begin{adjustwidth}{-0.3in}{-0.3in}
\centering
\begin{subfigure}{0.3\linewidth}
 \centering
 \includegraphics[width=\linewidth]{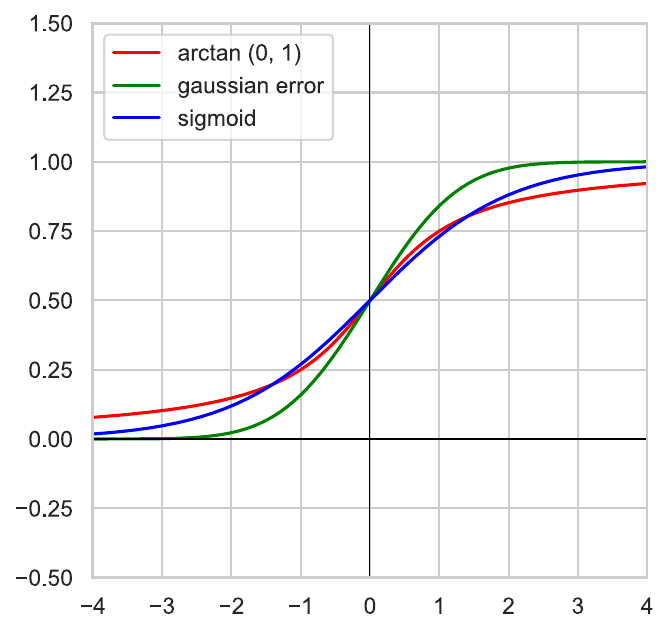}
 \caption*{(a)}
\end{subfigure}
\begin{subfigure}{0.3\linewidth}
 \centering
 \includegraphics[height=125pt, width=\linewidth]{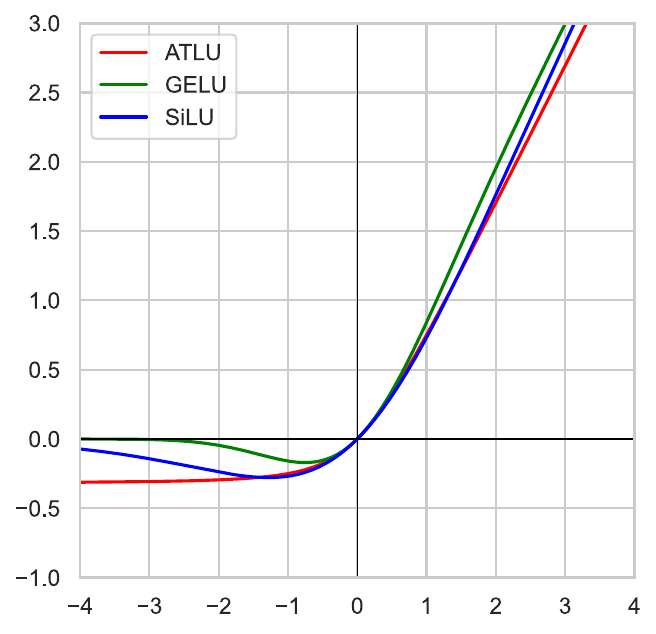}
 \caption*{(b)}
\end{subfigure}
\begin{subfigure}{0.3\linewidth}
 \centering
 \includegraphics[width=\linewidth]{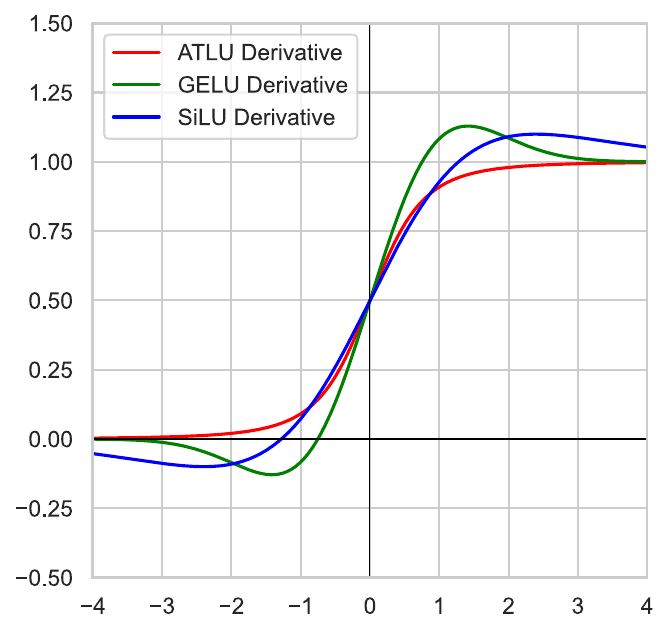}
 \caption*{(c)}
\end{subfigure}
\end{adjustwidth}
\caption{\textbf{Comparison of ATLU, GELU, and SiLU.} \textbf{(a)} Graph of gating functions for ATLU, GELU, and SiLU. All are continuously differentiable, monotonically increasing, and have a gating range of $(0, 1)$. \textbf{(b)} Graph of ATLU, GELU, and SiLU. ATLU differs from GELU and SiLU in that it is not ReLU-like: it does not converge to 0 as $x$ approaches negative infinity and does not converge to $x$ as $x$ approaches positive infinity. \textbf{(c)} Graph of first derivatives for ATLU, GELU, and SiLU. The first derivative of ATLU is monotonically increasing and does not have values below 0 or above 1, unlike GELU and SiLU.}
\label{fig:atlu-GELU-SiLU-comparison}
\end{figure}

The arctan function, also known as the inverse tangent, is continuously differentiable, monotonically increasing, and has a range of $(-\frac{{\pi}}{2}, \frac{{\pi}}{2})$. Given these properties, we explore the viability of using the arctan function as a gating mechanism. We define ArcTan Linear Unit (ATLU) as a self-gated activation function that uses the arctan function scaled to the range of (0,1) as its gating function:
\begin{equation}
\text{ATLU}(x) = x \times \frac{\text{arctan}(x) + \frac{{\pi}}{2}}{\pi}
\label{eq:5}
\end{equation}
A visualization of ATLU, its gating function, and its first derivative is provided in Figure \ref{fig:atlu-GELU-SiLU-comparison}. ATLU satisfies properties 1, 2, and 3 but does not satisfy properties 4 and 5, as listed in Section \ref{2.2-favorable-properties}. Unfortunately, ATLU empirically performs poorly compared to GELU and SiLU, and performs on par with ReLU. Therefore, we deduce that either converging to the same values as ReLU is important and/or allowing gradients below zero and above one is necessary for the effectiveness of an activation function.

\begin{figure}[h]
\begin{adjustwidth}{-0.3in}{-0.3in}
\centering
\begin{subfigure}{0.3\linewidth}
 \centering
 \includegraphics[height=125pt, width=\linewidth]{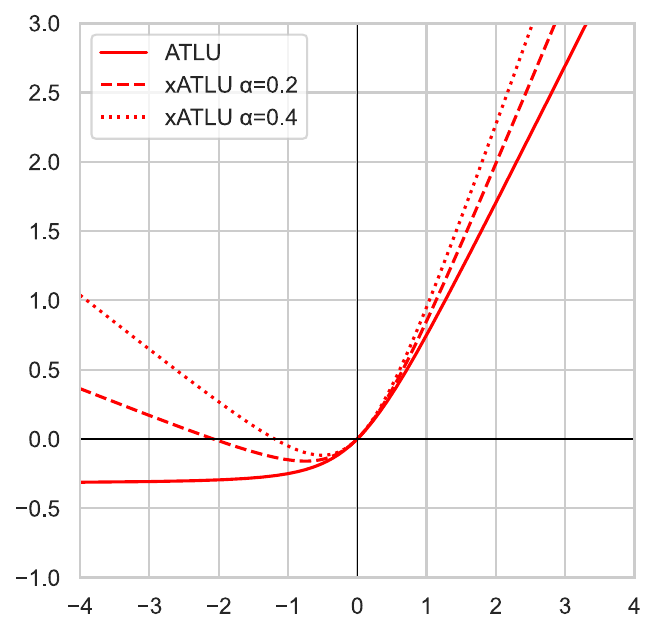}
 \caption*{(a)}
\end{subfigure}
\begin{subfigure}{0.3\linewidth}
 \centering
 \includegraphics[width=\linewidth]{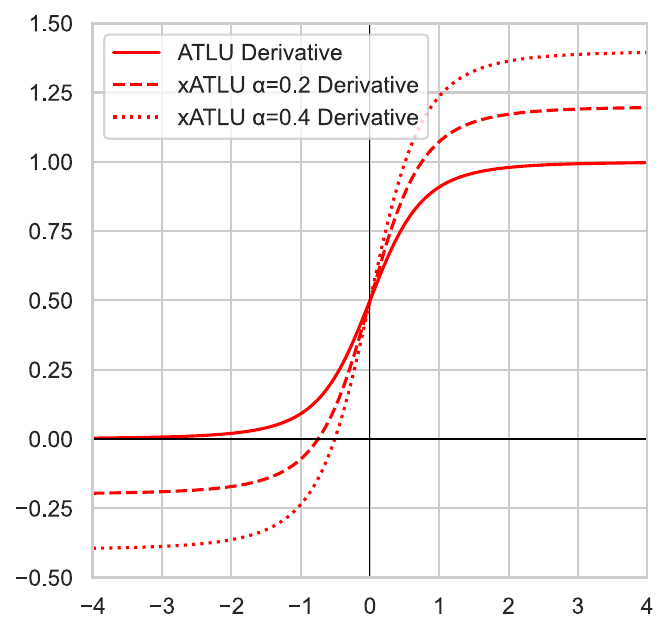}
 \caption*{(b)}
\end{subfigure}
\end{adjustwidth}
\caption{\textbf{Visualisation of xATLU.} (a) Graph of xATLU for various fixed values of $\alpha$. Increasing $\alpha$ makes the activation function converge towards more positive values in both directions. A similar effect occurs when applied to GELU and SiLU. (b) Graph of the first derivative of xATLU for various fixed values of $\alpha$. Increasing $\alpha$ increases the range of the first derivative. A similar effect occurs when applied to GELU and SiLU.}
\label{fig:expand-atlu}
\end{figure}

\subsection{Expanded Gating Ranges}

We modify ATLU to allow gradient values below zero and above one by expanding the range of $g(x)$. This is done by introducing a single trainable scalar $\alpha$, initialized to 0, for every MLP block that rescales the gating range to $(-\alpha, 1+\alpha)$. The rationale for this specific design choice is covered in model ablations in Section \ref{model-ablations}. We name this variant Expanded ATLU (xATLU) and express it as follows:
\begin{equation}
\text{xATLU}(x, \alpha) = x \times \left(\frac{\text{arctan}(x) + \frac{{\pi}}{2}}{\pi} \times (1 + 2 \times \alpha) - \alpha \right)
\label{eq:6}
\end{equation}
Note that while the expression can be further simplified, this form provides better clarity. The effect of different fixed values of $\alpha$ on the activation function is illustrated in Figure \ref{fig:expand-atlu}. This same idea can be applied to GELU and SiLU to obtain xGELU and xSiLU. We also test the effectiveness of this idea on $g(x)$ in the GLU setting.

\section{Results}

\begin{figure}[h]
\begin{adjustwidth}{-0.3in}{-0.3in}
\centering
\includegraphics[width=0.48\linewidth]{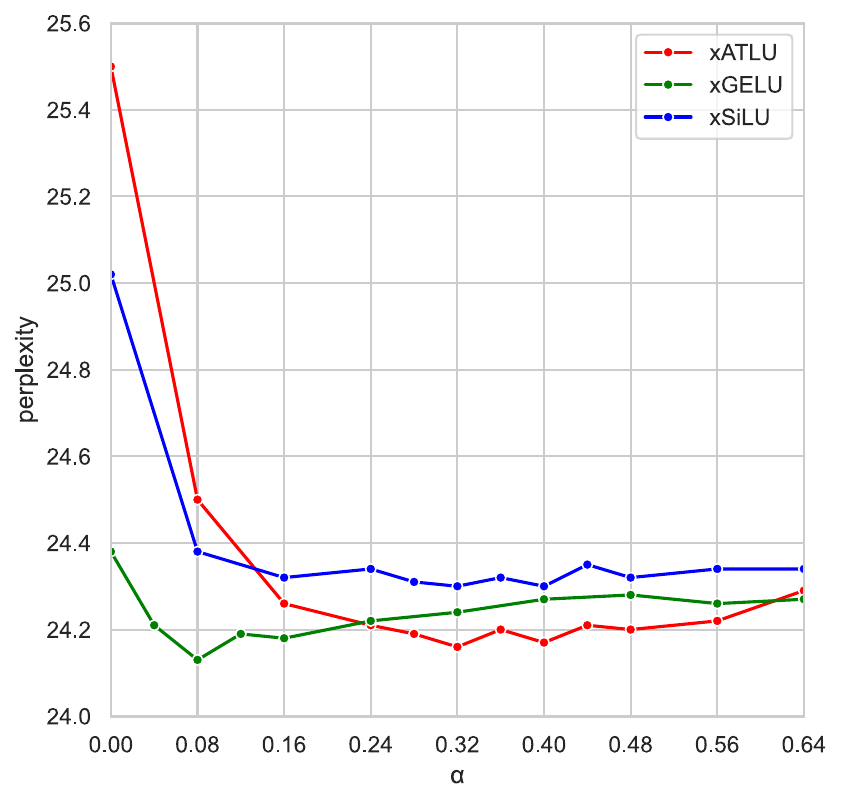}
\end{adjustwidth}
\caption{\textbf{Effect of Expanded Gating Ranges.} Experiments analysing the effect of using a fixed scalar value for $\alpha$. Note that using trainable scalar values performs better. The baseline activation functions ATLU, GELU and SiLU are at $\alpha=0$. Increasing $\alpha$ improves the performance of xATLU, xGELU and xSiLU, allowing them to surpass the performance of GELU and SiLU.}
\label{fig:expand-gating}
\end{figure}

We conduct experiments on standard transformer-based \citep{vaswani2017attention} autoregressive language modeling using code derived from Andrej Karpathy's nanoGPT implementation \citep{karpathy_nanoGPT_2022}. The experiments are run on single A100 GPUs on the OpenWebText2 dataset \citep{openwebtext2}, modifying only the activation function used within the MLP block. For standard MLP blocks, we use an MLP ratio of 4, and for gated MLP blocks, we use an MLP ratio of 8/3. We report the mean and standard error of the last 5 recorded perplexities over 3 different seeds. Full details on experiment setups are given in Appendix \ref{appendix:experiment-setup}.

\subsection{Self-Gated Activation Functions}

We first run small scale experiments to analyze the impact of fixed values of $\alpha$ for xATLU, xGELU, and xSiLU. The experiment setup is given in Appendix \ref{Effect of expanding gating ranges on xATLU, xGELU and xSiLU} and the results are visualized in Figure \ref{fig:expand-gating}. We make the following observations:

\textbf{Arctan is a Viable Gating Function.} ATLU is not a competitive activation function but expanded gating ranges allow xATLU to outperform GELU and SiLU.

\textbf{Expanded Gating Ranges Improve GELU and SiLU.} Expanded gating ranges allow xGELU and xSiLU to outperform GELU and SiLU.

\textbf{ReLU-like Properties are Not Necessary.} The performance of xATLU, xGELU, and xSiLU suggests that the optimal gating range is not between 0 and 1, and that the activation function does not need to converge to the same values as ReLU. We find that using a gating range of $(-\alpha, 1 + \alpha)$ improves performance.
 
\textbf{Negative Gradient Flow is Necessary.} Expanded gating ranges benefit xATLU the most, followed by xSiLU and xGELU. We theorize that there is an optimal amount of gradient flow, mainly negative as suggested by model ablations in Section \ref{model-ablations}, that $\alpha$ is responsible for controlling.
\begin{figure}[ht!]
\begin{adjustwidth}{-0.0in}{-0.0in}
\centering
\includegraphics[width=0.8\linewidth]{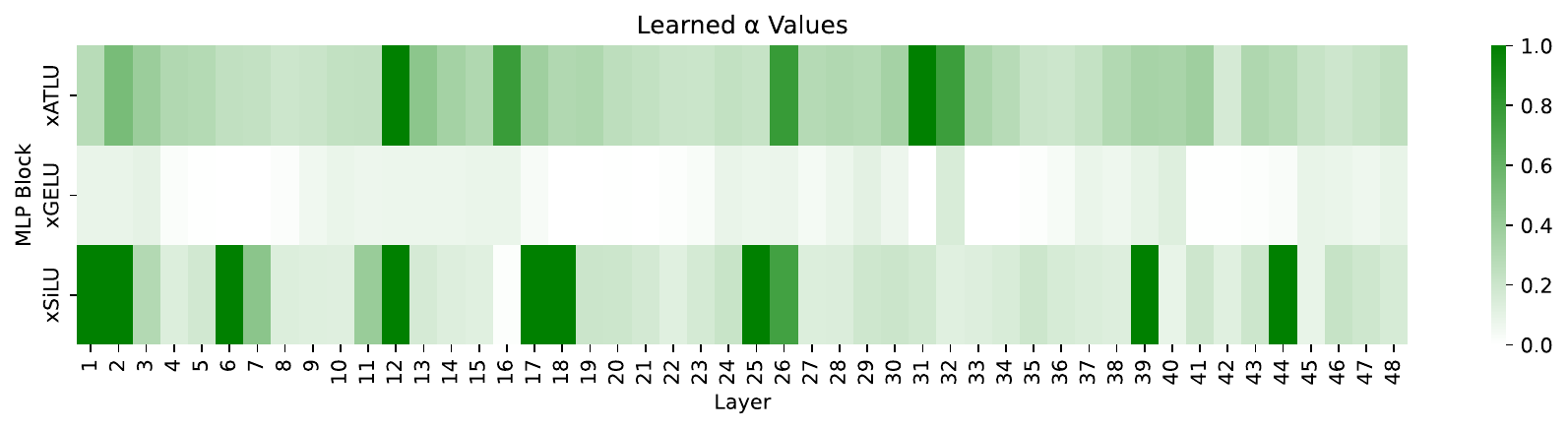}
\end{adjustwidth}
\caption{\textbf{Comparison of trainable $\alpha$ weights.} We use a heatmap to visualize learned $\alpha$ weights for depth 48 transformer models trained on OpenWebText2 using xATLU, xGELU and xSiLU. Note that we place no restrictions on the values that $\alpha$ can take and the learned values are all positive, which means $\alpha$ is expanding the gating ranges.}
\label{fig:alpha-visualisation}
\end{figure}
\begin{table}[]
\caption{\textbf{Performance of Self-Gated Activation Functions on OpenWebText2.} We report the activation function, the number of parameters, and the perplexity. xATLU, xGELU, and xSiLU outperform GELU and SiLU. xATLU outperforms xGELU and xSiLU.}
\label{tab:activation-functions-results}
\centering
\begin{tabular}{lcccc}
\midrule
$a(x)$ & $g(x)$ Range & Depth & \#Parameter (M) & Perplexity ($\downarrow$) \\
\midrule
ATLU & $(0, 1)$ & 12 & 124 & 18.13 $\pm$ 0.10 \\
GELU & $(0, 1)$ & 12 & 124 & 17.58 $\pm$ 0.10 \\
SiLU & $(0, 1)$ & 12 & 124 & 17.65 $\pm$ 0.10 \\
xATLU & $(-\alpha, 1 + \alpha)$ & 12 & 124 & \textbf{17.36 $\pm$ 0.10} \\
xGELU & $(-\alpha, 1 + \alpha)$ & 12 & 124 & 17.47 $\pm$ 0.10 \\
xSiLU & $(-\alpha, 1 + \alpha)$ & 12 & 124 & 17.46 $\pm$ 0.10 \\
\midrule
GELU & $(0, 1)$ & 24 & 209 & 16.05 $\pm$ 0.09 \\
xATLU & $(-\alpha, 1 + \alpha)$ & 24 & 209 & \textbf{15.71 $\pm$ 0.08} \\
xGELU & $(-\alpha, 1 + \alpha)$ & 24 & 209 & 15.79 $\pm$ 0.08 \\
xSiLU & $(-\alpha, 1 + \alpha)$ & 24 & 209 & 15.78 $\pm$ 0.09 \\
\midrule
GELU & $(0, 1)$ & 48 & 379 & 14.31 $\pm$ 0.08 \\
xATLU & $(-\alpha, 1 + \alpha)$ & 48 & 379 & \textbf{13.93 $\pm$ 0.08} \\
xGELU & $(-\alpha, 1 + \alpha)$ & 48 & 379 & 14.05 $\pm$ 0.08 \\
xSiLU & $(-\alpha, 1 + \alpha)$ & 48 & 379 & 14.00 $\pm$ 0.07 \\
\bottomrule
\end{tabular}
\end{table}

We run larger experiments using a trainable scalar for $\alpha$ for xATLU, xGELU and xSiLU. The experiment setup is given in Appendix \ref{Performance of Self-Gated Activation Functions}, the results are shown in Table \ref{tab:activation-functions-results}, and a visualization of trained $\alpha$ weights in Figure \ref{fig:alpha-visualisation}. We make the following observations:

\textbf{xATLU, xGELU, and xSiLU Outperform GELU and SiLU.} The experiments suggest that expanded gating ranges can improve existing activation functions.

\textbf{xATLU Outperforms xGELU and xSiLU.} We theorize that this is due to ATLU having a monotonically increasing first derivative, which results in more favorable training dynamics for xATLU. The first derivatives of GELU and SiLU have similar properties, leading to similar performance for xGELU and xSiLU.

\subsection{Gated Linear Units}

We run experiments to evaluate the effectiveness of expanded gating ranges for first and second-order GLU. Additionally, we explore the use of arctan scaled to the range (0, 1) in the GLU setting, which we name the ArcTan Gated Linear Unit (ATGLU). The experiment setup is given in Appendix \ref{Performance of Gated Linear Units}, and the results are shown in Table \ref{tab:glu-results}. We make the following observations:

\textbf{Expanded Gating Ranges Improve First-Order GLU} The results are similar to self-gated activation functions. First-order ATGLU has poor performance. First-order xATGLU, xGEGLU and xSwiGLU outperforms first-order ATGLU, GEGLU, and SwiGLU. First-order xATGLU outperforms first-order xGEGLU and xSwiGLU.

\textbf{Expanded Gating Ranges Do Not Improve Second-Order GLU.} Second-order ATGLU is competitive with second-order GEGLU and SwiGLU. This is surprising as arctan needed expanded gating ranges to function properly in the self-gated activation functions and first-order GLU setting. We theorize that both second-order GLU and expanded gating ranges achieve similar effects in facilitating a larger negative gradient flow.

\textbf{Expanded First-Order GLU Match Second-Order GLU.} Our results suggest that expanded gating ranges narrows the performance gap observed between first and second-order GLU. First-order xATGLU, xGEGLU, and xSwiGLU and second-order ATGLU, GEGLU and SwiGLU have similar performance.

\begin{table}[]
\caption{\textbf{Performance of Gated Linear Units on OpenWebText2.} We report the gated linear unit and order, the number of parameters, and the perplexity. We refer to GEGLU as second-order GEGLU, SwiGLU as second-order SwiGLU and the original GLU as first order SwiGLU. Expanded gating ranges benefits first-order GLU but not second-order GLU. First-order ATGLU, GEGLU, SwiGLU and second-order ATGLU appear to be able to match the performance of second-order GEGLU and SwiGLU.}
\label{tab:glu-results}
\centering
\begin{tabular}{lccccc}
\midrule
$a(x)$ & Order & $g(x)$ Range & Depth & \#Parameter (M) & Perplexity ($\downarrow$) \\
\midrule
ATGLU & 1st & $(0, 1)$ & 12 & 124 & 18.23 ± 0.15 \\
GEGLU & 1st & $(0, 1)$ & 12 & 124 & 17.84 ± 0.16 \\
SwiGLU & 1st & $(0, 1)$ & 12 & 124 & 17.85 ± 0.15 \\
xATGLU & 1st & $(-\alpha, 1 + \alpha)$ & 12 & 124 & \textbf{17.67 ± 0.15} \\
xGEGLU & 1st & $(-\alpha, 1 + \alpha)$ & 12 & 124 & 17.73 ± 0.16 \\
xSwiGLU & 1st & $(-\alpha, 1 + \alpha)$ & 12 & 124 & 17.69 ± 0.15 \\
\midrule
ATGLU & 2nd & $(0, 1)$ & 12 & 124 & 17.79 ± 0.15 \\
GEGLU & 2nd & $(0, 1)$ & 12 & 124 & \textbf{17.78 ± 0.16} \\
SwiGLU & 2nd & $(0, 1)$ & 12 & 124 & 17.81 ± 0.16 \\
xATGLU & 2nd & $(-\alpha, 1 + \alpha)$ & 12 & 124 & 17.82 ± 0.15 \\
xGEGLU & 2nd & $(-\alpha, 1 + \alpha)$ & 12 & 124 & 17.80 ± 0.16 \\
xSwiGLU & 2nd & $(-\alpha, 1 + \alpha)$ & 12 & 124 & 17.84 ± 0.16 \\
\midrule
xATGLU & 1st & $(-\alpha, 1 + \alpha)$ & 24 & 209 & \textbf{15.55 ± 0.09} \\
xGEGLU & 1st & $(-\alpha, 1 + \alpha)$ & 24 & 209 & 15.62 ± 0.09\\
xSwiGLU & 1st & $(-\alpha, 1 + \alpha)$ & 24 & 209 & 15.57 ± 0.09 \\
\midrule
ATGLU & 2nd & $(0, 1)$ & 24 & 209 & 15.60 ± 0.09 \\
GEGLU & 2nd & $(0, 1)$ & 24 & 209 & 15.58 ± 0.09 \\
SwiGLU & 2nd & $(0, 1)$ & 24 & 209 & \textbf{15.57 ± 0.09} \\
\bottomrule
\end{tabular}
\end{table}

\subsection{Model Ablations}
\label{model-ablations}

As expanded gating ranges have the largest impact on ATLU/xATLU, we focus on running model ablations for ATLU/xATLU. The experiment setup is given in Appendix \ref{Effect of expanding gating ranges on xATLU, xGELU and xSiLU} and the results are shown in Table \ref{tab:xatlu-ablation}. We make the following observations:

\textbf{Trainable Scalar Over Fixed Scalar.} Replacing the trainable scalar $\alpha$ with the best performing fixed scalar $\alpha$ from Figure \ref{fig:expand-gating} marginally worsens performance. However, it still significantly outperforms baseline ATLU.

\textbf{Importance of Negative Gradient Flow.} Replacing the gating range of $(-\alpha, 1 + \alpha)$ with $(0, 1 + \alpha)$ significantly worsens performance, $\alpha$ likely plays an important role in increasing the flow of negative gradients. Replacing it with $(-\alpha, 1)$ marginally worsens performance, suggesting that increasing the flow of positive gradients is less important.

\textbf{Alternative Parameterizations.} Replacing the gating range of $(-\alpha, 1 + \alpha)$ with $(-\alpha_{1}, 1 + \alpha_{2})$ results in similar performance. Replacing the scalar weight of $\alpha$ with a per-channel weight, which is model dimension multiplied by mlp ratio, also results in similar performance. We run our experiments with the most simple setup.

\begin{table}[]
\caption{\textbf{Ablation of xATLU.} We run model ablations using fixed scalar values for $\alpha$, using $\alpha$ to only expand the minimum gating or the maximum gating, using different parameters to control minimum gating and maximum gating, and replacing the trainable scalar weight with a trainable per-channel weight.}
\label{tab:xatlu-ablation}
\centering
\begin{tabular}{lcccc}
\midrule
$a(x)$ & $g(x)$ Range & Depth & \#Parameter (M) & Perplexity ($\downarrow$) \\
\midrule
ATLU & $(0, 1)$ & 12 & 124 & 18.13 ± 0.10 \\
xATLU & $(-\alpha, 1 + \alpha)$ & 12 & 124 & \textbf{17.36 ± 0.10} \\
\midrule
xATLU & $(-0.32, 1.32)$ & 12 & 124 & 17.47 ± 0.10 \\
\midrule
xATLU & $(-\alpha, 1)$ & 12 & 124 & 17.53 ± 0.11 \\
xATLU & $(0, \alpha)$ & 12 & 124 & 18.08 ± 0.10 \\
\midrule
xATLU & $(-\alpha_{1}, 1 + \alpha_{2})$ & 12 & 124 & 17.38 ± 0.10 \\
\midrule
xATLU & \begin{tabular}[c]{@{}c@{}}$(-\alpha, 1 + \alpha)$\\ per channel\end{tabular} & 12 & 124 & \textbf{17.36 ± 0.10} \\
\bottomrule
\end{tabular}
\end{table}

\newpage

\section{Limitations and Future Work}

\textbf{Running Larger Scale Experiments.} Due to limited computational resources, our experiments are relatively small in scale. Larger experiments are necessary to determine if xATLU, xGELU, and xSiLU can consistently outperform GELU and SiLU, and if first-order xATGLU, xGEGLU, xSwiGLU, and second-order ATGLU can compete with second-order GEGLU and SwiGLU.

\textbf{Impact on Activation Sparsity.} Activation sparsity \citep{mirzadeh2023ReLU, song2024prosparse} has been identified as a potentially important property for both computational and memory efficiency. Expanding gating ranges deviates from a ReLU-like activation function, which may make it more challenging to achieve activation sparsity through techniques like ReLUfication.

\textbf{Searching for New Gating Functions.} Expanded gating ranges increases the search space of viable gating functions. GELU, SiLU, and Mish \citep{misra2020mish} all have different expressions for their gating functions but result in similarly shaped activation functions and first derivatives. It is likely that there exists gating functions with completely different expressions to arctan, and have similar activation functions and first derivatives to ATLU.

\section{Related Work}

\textbf{ReLU-like Activation Functions.} The ReLU activation function gained popularity due to its simplicity and efficiency. Most subsequent work on activation functions has adopted several ReLU-like properties \citep{hendrycks2016gaussian, ramachandran2017searching, elfwing2017sigmoidweighted, misra2020mish}. Our findings suggest that expanded gating ranges can improve multiple activation functions, indicating that some previously considered desirable ReLU-like properties may not be necessary.

\textbf{Trainable Activation Functions.} Several prior works have proposed trainable or adaptable activation functions \citep{he2015delving, ramachandran2017searching, Apicella_2021}. However, they have limited effectiveness, and are not used over non-trainable activation functions. Our research suggests that introducing a trainable parameter to control the gating range can be beneficial for activation functions. 

\section{Conclusion}

This work aims to enhance the understanding of activation functions. We identified key properties that contribute to the effectiveness of activation functions and demonstrated the viability of using arctan as a gating function. By expanding the gating ranges, we showed that self-gated activation functions such as xATLU, xGELU, and xSiLU can outperform popular activation functions GELU and SiLU. Furthermore, we demonstrated that expanded gating ranges can also improve first-order GLU and help bridge the performance gap with second-order GLU. 

\newpage

\section{Acknowledgements}

I would like to thank Martin Jaggi for insightful comments and suggestions during the development of this work.

\bibliographystyle{plainnat}
\bibliography{neurips_2024}


\newpage
\appendix
\section{Appendix}

\subsection{Experiment Setup}
\label{appendix:experiment-setup}

\begin{table}[H]
\caption{\textbf{Shared Hyperparameters.}}
\label{table:hyperparameters}
\vskip 0.15in
\begin{center}
\begin{small}
\begin{sc}
\begin{tabular}{ll}
\midrule
\textnormal{Hyperparameter} & \textnormal{Value} \\
\midrule
\textnormal{Batch Size} & \textnormal{500}\\
\textnormal{Weight Decay} & \textnormal{0.1}\\
\textnormal{Optimizer} & \textnormal{adamw}\\
\textnormal{Beta 1} & \textnormal{0.9}\\
\textnormal{Beta 2} & \textnormal{0.95}\\
\textnormal{Scheduler} & \textnormal{cosine}\\
\textnormal{Warmup Percent} & \textnormal{0.02}\\
\textnormal{Mininum Learning Rate Ratio} & \textnormal{0.1}\\
\textnormal{Dropout} & \textnormal{0.0}\\
\textnormal{Head Dimension} & \textnormal{64}\\
\bottomrule
\end{tabular}
\end{sc}
\end{small}
\end{center}
\vskip -0.1in
\end{table}

\subsubsection{Effect of expanded gating ranges on xATLU, xGELU and xSiLU}
\label{Effect of expanding gating ranges on xATLU, xGELU and xSiLU}

\begin{table}[H]
\caption{\textbf{Effect of expanded gating ranges on xATLU, xGELU and xSiLU.} Small scale experiments using fixed scalar for $\alpha$ to test viability of expanding the gating range. Trained on 3.84B tokens.}
\label{table:Effect of expanding gating ranges on xATLU, xGELU and xSiLU}
\vskip 0.15in
\begin{center}
\begin{small}
\begin{sc}
\begin{tabular}{ll}
\midrule
\textnormal{Hyperparameter} & \textnormal{Value} \\
\midrule
\textnormal{Iterations} & \textnormal{15000}\\
\textnormal{Sequence Length} & \textnormal{512}\\
\textnormal{Model Dimension} & \textnormal{512}\\
\textnormal{Depth} & \textnormal{12}\\
\textnormal{Learning Rate} & \textnormal{0.002}\\
\textnormal{Gradient Clipping} & \textnormal{0.0}\\
\bottomrule
\end{tabular}
\end{sc}
\end{small}
\end{center}
\vskip -0.1in
\end{table}

\subsubsection{Performance of Self-Gated Activation Functions}
\label{Performance of Self-Gated Activation Functions}

\begin{table}[H]
\caption{\textbf{Performance of Self-Gated Activation Functions.} Larger scale experiments using trainable scalar for $\alpha$. Trained on 17.92B tokens. Depth 12 models use 0.002 learning rate and Depth 24/48 models use 0.001 learning rate.}
\label{table:Performance of Self-Gated Activation Functions}
\vskip 0.15in
\begin{center}
\begin{small}
\begin{sc}
\begin{tabular}{ll}
\midrule
\textnormal{Hyperparameter} & \textnormal{Value} \\
\midrule
\textnormal{Iterations} & \textnormal{40000}\\
\textnormal{Sequence Length} & \textnormal{768}\\
\textnormal{Model Dimension} & \textnormal{768}\\
\textnormal{Learning Rate} & \textnormal{0.002/0.001}\\
\textnormal{Gradient Clipping} & \textnormal{0.0}\\
\bottomrule
\end{tabular}
\end{sc}
\end{small}
\end{center}
\vskip -0.1in
\end{table}

\subsubsection{Performance of Gated Linear Units}
\label{Performance of Gated Linear Units}

\begin{table}[H]
\caption{\textbf{Performance of Gated Linear Units.} Larger scale experiments for GLU using trainable scalar for $\alpha$. Trained on 17.92B tokens. Depth 12 models use 0.002 learning rate and Depth 24 models use 0.001 learning rate. Uses 0.1 gradient clipping do reduce divergence during training due to the poorer training stability of GLU.}
\label{table:Performance of Gated Linear Units}
\vskip 0.15in
\begin{center}
\begin{small}
\begin{sc}
\begin{tabular}{ll}
\midrule
\textnormal{Hyperparameter} & \textnormal{Value} \\
\midrule
\textnormal{Iterations} & \textnormal{40000}\\
\textnormal{Sequence Length} & \textnormal{768}\\
\textnormal{Model Dimension} & \textnormal{768}\\
\textnormal{Learning Rate} & \textnormal{0.002/0.001}\\
\textnormal{Gradient Clipping} & \textnormal{0.1}\\
\bottomrule
\end{tabular}
\end{sc}
\end{small}
\end{center}
\vskip -0.1in
\end{table}

\subsection{Compute Resources Used}

For experiments with 40,000 iterations, sequence length 768, model dimension 768 and running on a single A100 GPU, 12 layer experiments take a day, 24 layer experiments take 2 days and 48 layer experiments take 4 days.

\subsection{ATLU/xATLU Derivatives}
\label{appendix:derivatives}

ATLU first derivative
\begin{equation}
\dfrac{\arctan\left(x\right)+\frac{{\pi}}{2}}{{\pi}}+\dfrac{x}{{\pi}\cdot\left(x^2+1\right)}
\end{equation}

xATLU first derivative
\begin{equation}
\alpha\left(\dfrac{\arctan\left(x\right)+\frac{{\pi}}{2}}{{\pi}}+\dfrac{x}{{\pi}\cdot\left(x^2+1\right)}\right) - \frac{\alpha}{2}
\end{equation}

\newpage

\section{Pseudocode}

\begin{verbatim}
import math
import torch
import torch.nn as nn

class xATLU(nn.Module):
    def __init__(self):
        super(xATLU, self).__init__()
        self.alpha = nn.Parameter(torch.zeros(1))
        self.half_pi = math.pi / 2
        self.inv_pi = 1 / math.pi

    def forward(self, x):
        gate = (torch.arctan(x) + self.half_pi) * self.inv_pi
        return x * (gate * (1 + 2 * self.alpha) - self.alpha)


class xGELU(nn.Module):
    def __init__(self):
        super(xGELU, self).__init__()
        self.alpha = nn.Parameter(torch.zeros(1))

    def forward(self, x):
        gate = (torch.erf(x / math.sqrt(2)) + 1) * 0.5
        return x * (gate * (1 + 2 * self.alpha) - self.alpha)


class xSiLU(nn.Module):
    def __init__(self):
        super(xSiLU, self).__init__()
        self.alpha = nn.Parameter(torch.zeros(1))

    def forward(self, x):
        gate = torch.sigmoid(x)
        return x * (gate * (1 + 2 * self.alpha) - self.alpha)
\end{verbatim}

\end{document}